\documentclass[]{fairmeta}
\usepackage{booktabs} % For formal tables
\usepackage{multirow}
\usepackage{enumitem}
\usepackage{makecell}
\usepackage{colortbl}
\usepackage{booktabs} % 用于 \toprule, \midrule, \bottomrule
\usepackage[table]{xcolor} % 用于 \rowcolor
\usepackage{graphicx}
\usepackage{algorithmic}
\usepackage[ruled]{algorithm2e} % For algorithms

\usepackage{amsmath,amsfonts,bm}

\def\eqref#1{equation~\ref{#1}}
\def\1{\bm{1}}

\def\mI{{\bm{I}}}

\def\mT{{\bm{T}}}

\DeclareMathAlphabet{\mathsfit}{\encodingdefault}{\sfdefault}{m}{sl}
\SetMathAlphabet{\mathsfit}{bold}{\encodingdefault}{\sfdefault}{bx}{n}
\newcommand{\tens}[1]{\bm{\mathsfit{#1}}}

\def\tB{{\tens{B}}}

\def\tF{{\tens{F}}}
\def\tG{{\tens{G}}}

\def\tI{{\tens{I}}}

\def\tV{{\tens{V}}}

\def\tX{{\tens{X}}}

\newcommand{\R}{\mathbb{R}}

\title{TransText: Alpha-as-RGB Representation for Transparent Text Animation}

\author[1,2,3*]{Fei Zhang}
\author[3]{Zijian Zhou}
\author[1,2]{Bohao Tang}
\author[3]{Sen He}
\author[3]{Hang Li}
\author[3]{Zhe Wang}
\author[3]{Soubhik Sanyal}
\author[1,2]{Pengfei Liu}
\author[3]{Viktar Atliha}
\author[3]{Tao Xiang}
\author[3]{Frost Xu}
\author[3]{Semih Gunel}

\affiliation[1]{Shanghai Jiao Tong University}
\affiliation[2]{Shanghai Innovation Institute (SII)}
\affiliation[3]{Meta AI}

\contribution[*]{Work done at Meta}
\abstract{We introduce the first method, to the best of our knowledge, for adapting \emph{image-to-video} models to layer-aware text (glyph) animation, a capability critical for practical dynamic visual design.
Existing approaches predominantly handle the transparency-encoding ($\alpha$ channel) as an extra latent dimension appended to the RGB space, necessitating the reconstruction of the underlying RGB-centric \emph{variational autoencoder} (VAE).
However, given the scarcity of high-quality transparent glyph data, retraining the VAE is computationally expensive and may erode the robust semantic priors learned from massive RGB corpora, potentially leading to latent pattern mixing.
To mitigate these limitations, we propose \textbf{TransText}, a framework based on a novel \emph{Alpha-as-RGB} paradigm to jointly model appearance and transparency without modifying the pre-trained generative manifold. 
\textbf{TransText} embeds the $\alpha$ channel as an RGB-compatible visual signal through latent spatial concatenation,  
explicitly ensuring strict cross-modal (RGB-and-Alpha) consistency while preventing feature entanglement. Our experiments demonstrate that \textbf{TransText} significantly outperforms baselines, generating coherent, high-fidelity transparent animations with diverse, fine-grained effects.}

\date{\today}
\metadata[Website]{\url{https://sii-ferenas.github.io/TransText}}
\begin{document}

\maketitle

\begin{figure}
  \centering
  % 宽度通常设为 \textwidth 以占满整行
  \includegraphics[width=\textwidth]{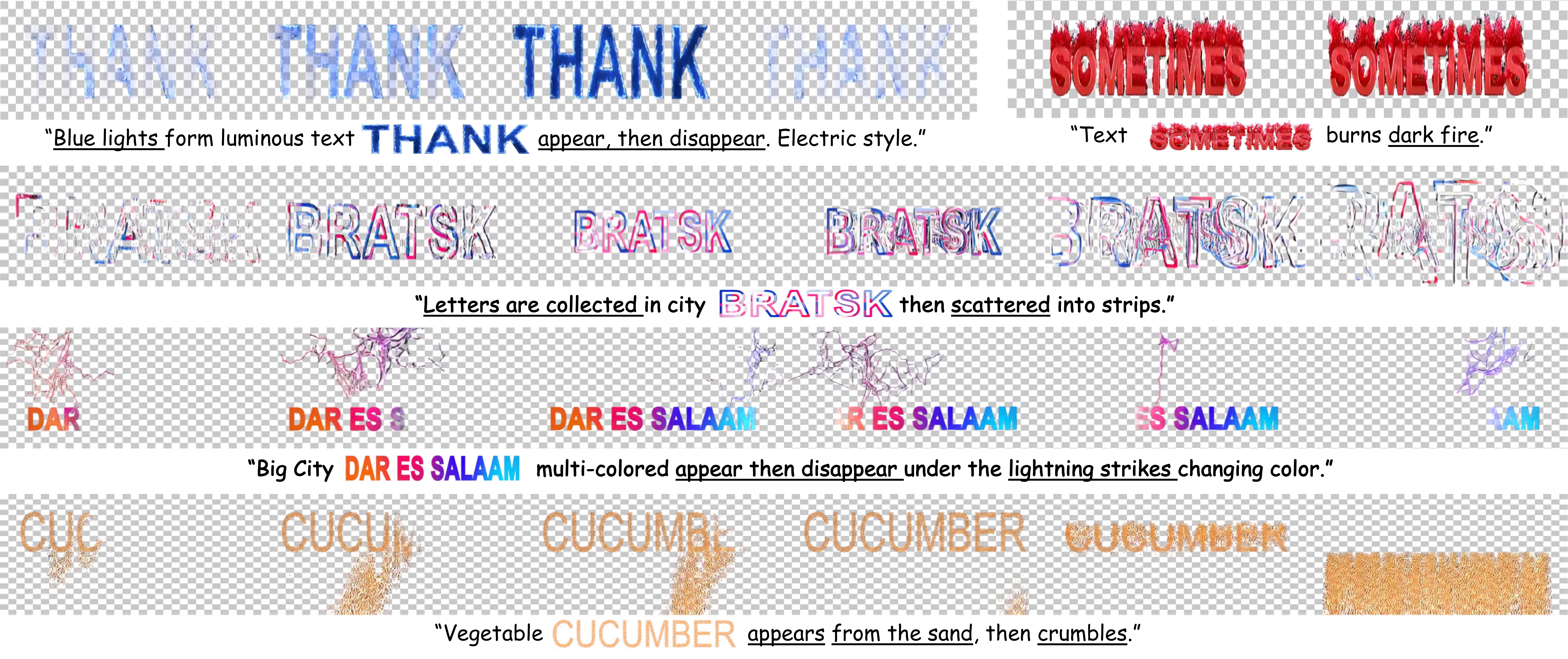} 
  \caption{\textbf{Visual frame-wise results of our \emph{image-to-video} (I2V) RGBA glyph animation model.} TransText generates \underline{diverse and complex} glyph animations with accurate, clean transparency while strictly preserving the style of the reference image. See the Supplementary Materials for additional video results.}
  \label{fig_teaser}
\end{figure}

\section{Introduction}

Recent advancements in diffusion models have revolutionized video generation, enabling high-quality synthesis across various modalities~\cite{blattmann2023stable,xu2024easyanimate,liu2024sora,yang2024cogvideox,wan2025wan}. While mainstream research predominantly targets standard RGB videos, attention is increasingly shifting towards \emph{RGB-Alpha} (RGBA) generation—a format that empowers creators with transparent assets for effortless compositing. Among potential assets, glyphs are indispensable in daily visual communication and professional workflows. To this end, targeting this specific yet widely applicable domain, this paper presents the first method dedicated to
RGBA video generation, effectively filling the gap in generating transparent, motion-rich animated textual content.

High-fidelity synthesis requires precise $\alpha$-channel modeling to ensure RGB-Alpha alignment. 
Most prior approaches~\cite{zhang2024transparent,huang2024layerdiff,kang2025layeringdiff,chen2025transanimate,yang2025layeranimate,dong2025wan,yin2025qwen} treat $\alpha$ as an intrinsic component of the latent space. 
This requires reconstructing and retraining the underlying \emph{variational autoencoder} (VAE) to support $\alpha$-aware encoding which entails a critical limitation. 
Given the scarcity of high-quality RGBA glyph data,
retraining billion-scale RGB-tuned VAEs~\cite{rombach2022high,wan2025wan} under these conditions incurs substantial computational cost and risks degrading the semantic priors learned from large-scale RGB corpora.

These limitations drives us to pose the question: \emph{Can Alpha be learnt directly as RGB?}
Recent advancements~\cite{wang2023images,bai2024sequential} demonstrate that diverse visual signals, e.g., semantic segmentation maps and depth fields, can be effectively modeled by encoding them into RGB-compatible formats, facilitating a unified training paradigm.
And fundamentally, the $\alpha$-channel shares the same spatial-temporal nature as the RGB visual signals.
Motivated by this, we propose to transform the single-channel Alpha into an RGB-like representation via channel replication, thereby establishing a unified format for both RGB and Alpha streams.
This \emph{Alpha-as-RGB} strategy allows us to treat transparency as a standard video sequence, facilitating RGB-Alpha latent co-learning for the pre-trained RGB foundation model without re-training the VAE.

To instantiate this \emph{Alpha-as-RGB} strategy, we propose \textbf{TransText}, an end-to-end framework tailored for image-supported glyph RGBA animation.  We introduce a latent \emph{spatial alignment} mechanism that processes Alpha and RGB streams with high spatio-temporal consistency. Specifically, we concatenate Alpha and RGB latents along the spatial dimension rather than the temporal dimension. This treats both components as a cohesive video sequence while isolating their feature patterns. Spatial concatenation prevents cross-modal entanglement. In contrast, temporal stacking causes the model to conflate RGB texture dynamics with transparency evolution.
Moreover, glyph transparency requires higher structural precision than RGB textures. Standard diffusion targets (e.g., velocity) often fail to preserve high-frequency geometric details as they deviate from the natural data manifold~\cite{li2025back}. 
To relieve this, we propose a fine-grained regularization term for the Alpha latent. This objective supervises the one-step denoised latent against the original input. It anchors generation to the data manifold, enforcing structural sharpness and stabilizing transparency boundaries.
Overall, we make the following contributions:

\begin{itemize}[leftmargin=*]
  \item   We explore the potential of the \emph{Alpha-as-RGB} representation for joint RGB and transparency learning in glyph animation. This eliminates costly VAE retraining and enables direct adaptation of off-the-shelf diffusion models to transparent video generation.

  \item We propose \textbf{TransText}, the first \emph{image-to-video} framework to implement the \emph{Alpha-as-RGB} representation for RGBA glyph animation. It enforces RGB-$\alpha$ consistency through spatial alignment and refines transparency via latent-level regularization.
  
  \item Experiments across diverse typographic effects show that \textbf{TransText} outperforms existing methods. It achieves \textbf{-453.53} FVD and \textbf{+29.40} Soft $\alpha$-mIoU improvement on average, demonstrating robust and high-fidelity transparent glyph animation.
\end{itemize}

\section{Related Work}\label{sec_related_work}
\noindent\textbf{Video Generation \& Text Animation.}
Recent advances in diffusion models~\cite{ho2020denoising, dhariwal2021diffusion} have led to remarkable progress in video generation. Early work~\cite{singer2022make, ho2022imagen, guo2023animatediff, xu2024easyanimate, xing2024make} has primarily focused on tailoring powerful text-to-image-based diffusion models~\cite{ho2022video, rombach2022high, ramesh2022hierarchical} by incorporating motion-aware modules and leveraging video datasets to achieve temporally coherent results. Furthermore, the emergence of highly-scalable diffusion transformers~\cite{peebles2023scalable} has enabled the development of large-scale video foundation models~\cite{yang2024cogvideox, wan2025wan},  which serve as the standard pipeline for customizing various downstream tasks. Beyond the \emph{text-to-video} (T2V) paradigm, some studies have also explored \emph{image-to-video} (I2V) frameworks~\cite{blattmann2023stable, chen2023seine, chen2023videocrafter1, xing2024dynamicrafter, wan2025wan}, by introducing effective latent fusion mechanisms between CLIP-encoded~\cite{clip} image information and noised latent representations, thereby enabling more flexible and controllable video generation.

Considering the substantial practical and economic value of glyphs (e.g., advertisements, posters, and documents), increasing attention has been devoted to visual text generation. In general, most of them~\cite{chen2023textdiffuser, tuo2023anytext, zhu2024visual, chen2024textdiffuser, tuo2024anytext2, liu2024glyph, shi2025fonts} focus on image-centered scene text rendering following similar technological pathways to those used in the natural object domain. Specifically, they turned to fine-tuning off-the-shelf diffusion models~\cite{rombach2022high} in a parameter-efficient manner~\cite{hu2022lora}. The unique fine-grained variations of text (e.g., color, font, and layout) have led to diverse emphases in glyph generation research. For instance,~\citet{chen2023textdiffuser, zhu2024visual} proposed to use character-level segmentation masks~\cite{zhang2025context} during the noise prediction stage to achieve high-fidelity and accurate shape generation.  In contrast, progress in glyph animation/video generation has not been as substantial and remains underexplored. Early attempts~\cite{wong1996temporal, lee2002kinetic, kato2015textalive} mainly centered on developing GUI-like tools for integrating static typography with simple movement. Subsequently, \citet{men2019dyntypo, yang2021shape} proposed style-transfer models that animate static letters using styles extracted from examples, yet these approaches are limited by fixed glyph positions. Building upon the diffusion-based paradigm,~\citet{park2024kinetic} introduced the first work, following~\citet{guo2023animatediff}, that incorporates an image diffusion model with motion modules for text animation. Despite its effectiveness, the generated videos often suffer from limited temporal length and consistency due to the image-based model as the starting point. Moreover, this text-conditioned paradigm is limited in conveying complex user intent. To address these issues, this work aims to build the first text animation framework based on a powerful I2V foundation model, thereby achieving promising and high-fidelity text video results.

% ~\cite{zhang2023text2layer,zhang2024transparent,huang2024layerdiff,wang2025alphavae,huang2025dreamlayer,ji2025layerflow,huang2025psdiffusion,kang2025layeringdiff,yang2025layeranimate,wang2025transpixeler,chen2025transanimate} 
\noindent\textbf{Generation with Transparency.} Beyond RGB-centric generation, a line of works have been dedicated to a more challenging setting that generates layered or transparent content (i.e., with explicit Alpha-channel support) for regions of interest, enabling more flexible, product-grade compositing and editing. Intuitively, a key challenge lies in modeling the Alpha-channel representation, which serves as the foundation for high-fidelity RGBA generation. To this end, \cite{zhang2024transparent} pioneered a two-stage framework: first, extending the VAE to encode $\alpha$-aware latents; second, conducting diffusion in the joint RGBA latent space. This paradigm has since become the \emph{de facto} standard for RGBA image synthesis, with most follow-ups improving multi-layer realism via advanced latent fusion mechanisms~\cite{zhang2023text2layer,huang2024layerdiff,huang2025dreamlayer,huang2025psdiffusion,kang2025layeringdiff,dong2025wan,chen2025transanimate}. In contrast, \cite{wang2025transpixeler,ji2025layerflow} turned to direct concatenation between RGB and Alpha features for joint noise prediction, efficiently bypassing explicit $\alpha$-representation learning. Building upon this, this work develops the first framework for RGBA image-to-video glyph rendering in an $\alpha$-free learning paradigm.

\noindent\textbf{Video Matting.}Estimating the 
$\alpha$-matte for object-level transparency has been broadly explored in computer vision~\cite{zhang2021attention,lin2021real,li2024vmformer}, with methods largely bifurcating by trimap dependency. Trimap-based approaches \cite{zhang2021attention} exploit spatial-temporal feature aggregation to enforce temporal consistency. Trimap-free methods, in contrast, aim to relieve the annotation burden, thereby streamlining the matting pipeline. Early works \cite{lin2021real,sengupta2020background} rely solely on a first-frame background, while later efforts integrate segmentation priors for robustness in unconstrained scenes \cite{lin2022robust}. Most recently, end-to-end frameworks have been proposed using diverse architectures \cite{li2023videomatt,li2024vmformer}. Beyond task-specific designs, \cite{zhang2025context} presents an in-context, multi-task text recognition model readily extensible to matting. Motivated by these advances—and synergies with video generation models, this paper adopts a generation-then-prediction paradigm to construct a straightforward RGBA video generation baseline based on \cite{zhang2025context}, thereby empirically demonstrating its inherent limitations.
\section{Preliminaries}\label{sec_prelim}
\noindent\textbf{Visual Transparency Representation.} The visual transparency normally is expressed via an extra channel beyond RGB space. Formally, given a $f$-frame video sequence $\tV = \{\tI_{i} \in \R^{3 \times h \times w}\}^{f}_{i=1}$, where each image frame $\tI_{i}$, with a height of $h$ and width of $w$, is treated as a linear composition of a foreground image $\tF_{i} \in \R^{3 \times h \times w}$ and a background image $\tB_{i} \in \R^{3 \times h \times w}$ blended with $\alpha_{i} \in \R^{1 \times h \times w}$:
{
\begin{equation}
    \tI_{i} = \alpha_{i} \odot  \tF_{i} + (1-\alpha_{i}) \odot 
 \tB_{i}, \quad \alpha_{i} \in [0,1],
\end{equation}
}
where $\odot$ indicates element-wise multiplication, and $\alpha_{i}$ denotes the Alpha matte map. Unlike natural images with multiple semantic layers, glyph generation typically features a single foreground layer (text). Accordingly, this paper focuses on the single foreground layer with transparency setting. Thus, we set $\tB$ to a uniform black (0-valued) background, so that $\tI_{i} = \alpha_{i} \odot \tF_{i}$ for simplicity.  

\noindent\textbf{Diffusion-based Video Generation Pipeline.} Prevailing video generation pipelines~\cite{yang2024cogvideox,wan2025wan} predominantly adopt a continuous-time denoising process formalized via flow matching~\cite{lipman2022flow, esser2024scaling}. In this work, we build upon the I2V framework of Wan~\cite{wan2025wan}. Formally, an input video $\tV$ is first encoded into a compressed latent representation using a pretrained \emph{variational autoencoder} (VAE): $\mathcal{E}(\tV) = \tX^{1} \in \R^{f/4 \times h/8 \times w/8}$. Subsequently, we employ flow matching to learn a continuous-time generative diffusion process, which models the mapping between the target latent $\tX = \tX^{t}$ and a random Gaussian noise 
 $\tX^{0} \sim \mathcal{N}(0, I)$ via a fixed, linearly sampled timestep process: $\tX^{t} = t\tX^{1} + (1-t)\tX^{0}, t \in [0,1]$. In this way, a $\theta$-parameterized velocity prediction model $\mathcal{V}: \R^{f/4 \times h/8 \times w/8} \rightarrow \R^{f/4 \times h/8 \times w/8}$, instantiated from \emph{diffusion transformer} (DiT) is trained using the following \emph{mean squared error} (MSE) objective:
\begin{equation}
    \mathcal{L}_{\mathrm{mse}} = \mathbb{E}_{\tX^{0}, \tX^{1}, \mT_\mathrm{ref}, \mI_{\mathrm{ref}}}||\mathcal{V} (\tX^{t},\mT_\mathrm{ref}, \mI_{\mathrm{ref}};\theta) - d\tX^{t}/dt||_2^2,
\end{equation}
where the target velocity is given by $d\tX^{t}/dt = \tX^{1} - \tX^{0}$. Here $\mT_\mathrm{ref}$ and $\mI_\mathrm{ref}$ denote text and image embeddings extracted by off-the-shelf encoders (umT5~\cite{raffel2020exploring} and CLIP~\cite{clip}, respectively). These embeddings are fused with the original latent representation via cross-attention, providing conditional reference for controllable video generation. During inference, a new video can be readily generated by iteratively denoising a randomly sampled noise using 
$\mathcal{V}$, conditioned on the reference text and image.

\begin{figure*}[t]
    \centering
\includegraphics[width=0.9\linewidth, height=0.55\textheight, keepaspectratio]{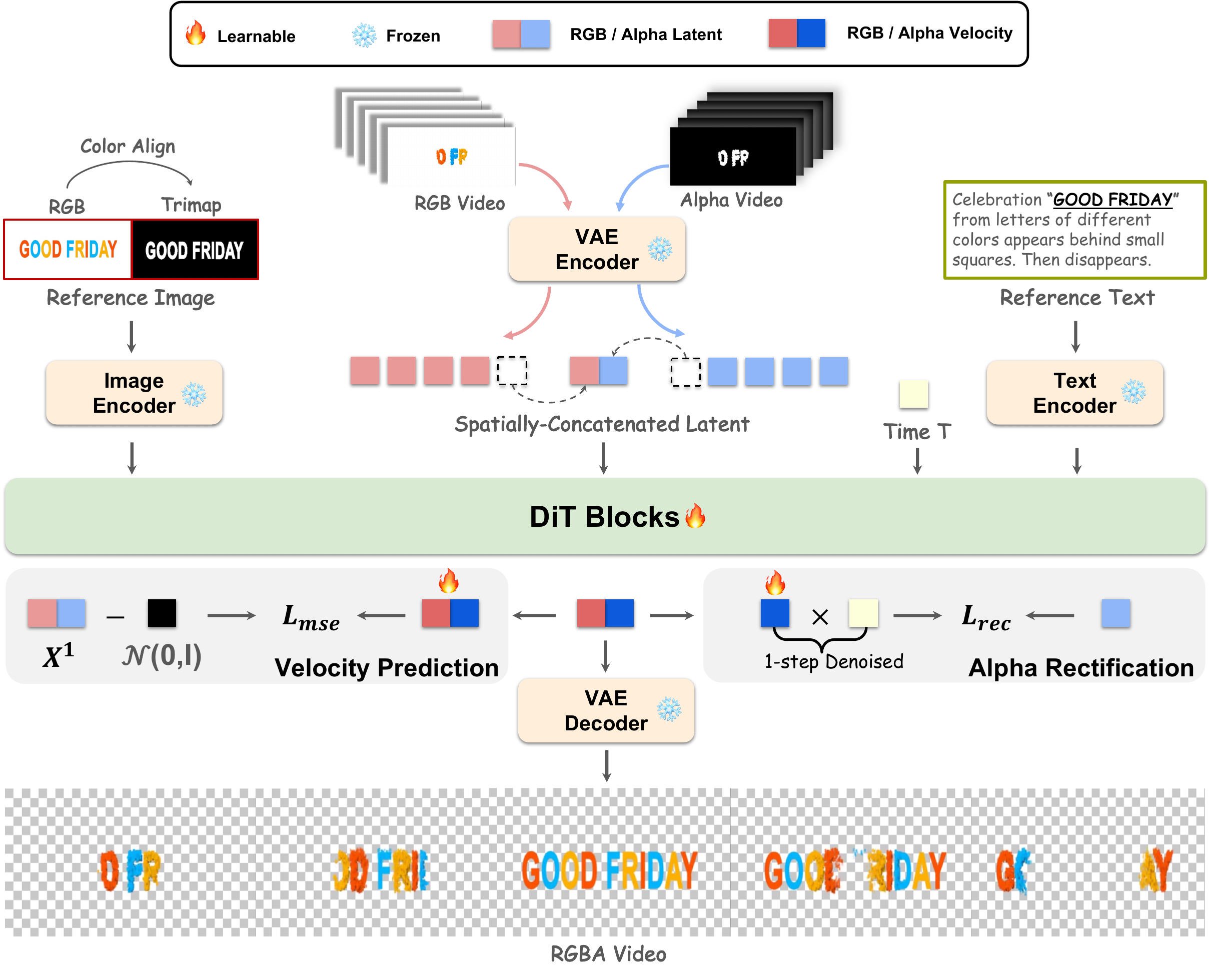} 
    \vspace{-2mm}
    \caption{\textbf{Overview of the \textbf{TransText} pipeline}. We obtain input latents by encoding the RGB video and the RGB-projected Alpha video (\emph{Alpha-as-RGB}) via the VAE. These latents are spatially concatenated for alignment. Additionally, the reference image and its derived trimap serve as structural conditions to guide the joint generation of both RGB textures and $\alpha$ mattes. During training, in addition to the standard velocity prediction loss $\mathcal{L}_{\mathrm{mse}}$, we introduce an $\alpha$-oriented  reconstruction term $\mathcal{L}_{\mathrm{rec}}$. Reconstruction loss performs one-step denoising using the predicted velocity to reconstruct the clean latent state, and explicitly aligns the reconstructed $\alpha$ with the ground-truth matte, thereby significantly improving fine-grained transparency generation.}
    \label{fig_main_fig}
    % \vspace{-2mm}
\end{figure*}

\section{Method}\label{sec_method}

Our method adapts an I2V generalist to generate per-frame transparent layers in a simple yet effective manner. As shown in Figure \ref{fig_main_fig}, \textbf{TransText} can generate transparent text animation with given reference text and image. Different from other VAE-training-based methods that have seperate RGB and Alpha representation  ~\cite{zhang2024transparent,dong2025wan},
we propose a unified, joint RGBA representation by explicitly representing the Alpha channel within the RGB space (\emph{Alpha-as-RGB}), thereby eliminating the need for additional latent encoder or special token learning. 

\subsection{Explicit Visual Alignment} 
\noindent\textbf{Transparency as Video.} Proper representation of Alpha is crucial for adapting conventional RGB-centric video foundation models to transparent content. ~\cite{zhang2024transparent,dong2025wan} address this by retraining a dedicated VAE that processes RGB and Alpha separately, which introduces significant training overhead and complicates joint learning. To avoid this burden, we turn to a VAE-training-free strategy that treats the Alpha matte as a standard RGB-compatible input. Specifically, we replicate the single-channel Alpha mask $\alpha_{i}$ across all three color channels, yielding a three-channel grayscale image. This allows us to construct an
\emph{Alpha-video} as $\tV_{\alpha} = \{ \tilde{\alpha}_{i} \in \R^{3 \times h \times w}\}^{f}_{i=1}$, where $\tilde{\alpha}_{i} = \texttt{round}(\alpha_{i} * 255) $ denotes the quantized, RGB-formatted matte frame. In this way,  both RGB foreground $\tV$ and $\tV_{\alpha}$ can be seamlessly processed by off-the-shelf video generation pipelines without architectural modification. Here we denote frames of alpha-video $\tV_{\alpha}$ as $\alpha_{i} = \tilde{\alpha}_{i} \in \R^{3 \times h \times w}$ by default.

\noindent\textbf{Spatial \emph{vs.} Temporal Alignment.} To enable joint training between $\tV$ and $\tV_{\alpha}$, concatenation in the latent space provides a straightforward way to link RGB $\tI$ and its transparency layer $\alpha$. Concatenation can be implemented in two ways based on the concatenation axis: \emph{spatial} (height- or width-wise) or \emph{temporal} (time-wise) alignment. Formally, spatial alignment concatenates each RGB image $\tI_i$ and corresponding transparency layer $\alpha_i$ map along the spatial dimension (e.g., width), yielding a spatially extended video latent $\tilde{\tX}_{\mathrm{spa}} =  [\mathcal{E}(\tV), \mathcal{E}(\tV_\mathrm{\alpha})] \in \R^{f/4 \times h/8 \times w/4}$ (a height-wise variant follows analogously). In contrast, temporal alignment doubles the video latent length by stacking $\mathcal{E}(\tV)$ and $\mathcal{E}(\tV_\mathrm{\alpha})$ temporally ($f$-dimension). Here we adopt spatial alignment for two key advantages::

% resulting in $\tV_{\mathrm{tem}} = [\tV, \tV_{\alpha}] \in \R^{2f \times 3 \times h \times w}$.

\textcircled{1} \textbf{Better Reference-oriented Adaptation}: With spatial alignment, the reference image can be formed by simple duplication along the same spatial axis. In this way, we adopt \emph{conditional color alignment} to further enhance guidance for $\alpha$ generation in the duplicated region, which maps the copied reference image into a binary-scale representation $\tG_{\mathrm{ref}} \in \R^{3 \times h \times w}$ as
\begin{equation}\small
    \tG_{\mathrm{ref}}^j = 
    \begin{cases}
  [0,0,0] & {\tI_{\mathrm{ref}}}^j < \beta \\
  [255, 255,255]  & {\tI_{\mathrm{ref}}}^j \geq \beta
\end{cases}
\quad, j \in [0, h * w]
\end{equation}
where $\beta$ is a threshold that binarizes the content. Intuitively, $\tG_{\mathrm{ref}}$ functions as a trimap, yielding a composite reference image compatible with spatial alignment: $\tilde{\tI}_{\mathrm{ref}} = [\tI_{\mathrm{ref}},\tG_{\mathrm{ref}}] \in \R^{3 \times h \times 2w}$. 
However, for  $\mathbf{V}_{\mathrm{tem}}$, the reference image can only be represented by either the original RGB foreground $\mathbf{I}_{\mathrm{ref}}$ or the $\alpha$  matte $\mathbf{G}_{\mathrm{ref}}$, due to the inherent limitation that static images cannot be temporally concatenated. Consequently, the model lacks explicit guidance to disentangle the RGB and $\alpha$ channels during generation, often resulting in confused learning dynamics where $\alpha$ erroneously inherits texture or semantic patterns from the RGB input, e.g., color information leaking into the $\alpha$ matte, which is empirically analyzed in 
Section~\ref{sec_ablation_study}.

\textcircled{2} \textbf{Better RGB-Alpha Alignment}: As discussed in~\cite{lu2024freelong,cai2025mixture,wan2025wan}, extended sequence lengths significantly impede convergence rates and stability. Temporal concatenation doubles the sequence length, exacerbating these optimization challenges. Crucially, learning such extended temporal dependencies is inherently data-hungry. Given the scarcity of high-quality transparent glyph data, the model struggles to robustly generalize over these doubled sequences, resulting in RGB-Alpha motion inconsistency (as shown in Table~\ref{app_tab_ablation}). In contrast, our spatial concatenation preserves the original temporal duration consistent with~\cite{wan2025wan}, avoiding the pitfalls of long-sequence modeling and ensuring efficient convergence even under data-constrained conditions.

\subsection{Fine-grained Transparency Rectification}
Alpha prediction poses inherent challenges when implemented within a generative pipeline~\cite{amit2021segdiff,chen2023diffusiondet,mousakhan2024anomaly}. Generative frameworks are naturally less suited for precise latent-level recognition—particularly in glyph-sensitive applications~\cite{chen2023textdiffuser,tuo2024anytext2}—and further suffer from an intrinsic limitation in synthesizing the full dynamic range required for $\alpha$ mattes. As noted by~\cite{guttenberg2023diffusion}, standard diffusion models struggle to alter the global mean (i.e., long-wavelength features) from zero-mean noise, making it difficult to generate the pure transparency or opacity values essential for sharp glyph boundaries. To address this limitation, we introduce an additional reconstruction regularization to refine the generated $\alpha$ latent. Our motivation stems from the insight that standard diffusion targets (like noise or velocity) often struggle in high-dimensional spaces due to their deviation from the natural data distribution, and~\cite{li2025back} demonstrated that directly predicting clean data allows models to operate effectively on the low-dimensional data manifold. Inspired by this, we leverage the reconstruction loss to guide the velocity learning process, forcing the model to respect the intrinsic geometric structure of the $\alpha$ channel rather than solely fitting the diffusion trajectory. Specifically, given the predicted alpha-component velocity $\mathcal{V}_{\alpha}$ at time $t$, we obtain an estimate of the target ($t=1$) latent via a single-step Euler integration:
\begin{equation}\small
    \tilde{\mathbf{X}}^{1}_{\alpha} = \mathbf{X}^{t}_{\alpha} + (1 - t) \, \mathcal{V}_{\alpha}(\mathbf{X}^{t}),
\end{equation}
where $\mathbf{X}^{t}$ denotes the full latent at time $t$. We then enforce latent-level alignment between this reconstruction and the ground-truth Alpha latent $\mathbf{X}^{1}_{\alpha}$ using an MSE loss:
\begin{equation}\small
    \mathcal{L}_{\mathrm{rec}} = \left\| \mathbf{X}^{1}_{\alpha} - \tilde{\mathbf{X}}^{1}_{\alpha} \right\|_2^2.
\end{equation}

The final training objective for the velocity prediction network $\mathcal{V}$ combines the standard flow-matching (or MSE) loss with the following reconstruction term:
\begin{equation}\small
    \mathcal{L} = \mathcal{L}_{\mathrm{mse}} + \lambda \, \mathcal{L}_{\mathrm{rec}},
\end{equation}
where $\lambda$ is a fixed hyperparameter balancing the two components. 

\section{Experiment}\label{sec_experiment}
\subsection{Experimental Settings}\label{sec_exp_setting}
\noindent{\textbf{Benchmark \& Evaluation Metric}}. Due to the scarcity of publicly available text RGBA animation benchmarks, we have curated a high-quality and high-fidelity text animation dataset by collecting samples from \textit{ShutterStock} \footnote{https://www.shutterstock.com/}, a large corpus of online image and video resources. After thorough annotation and filtering, we constructed a dataset comprising a total of 16,108 valid RGBA text animation samples, spanning 8 distinct visual effects. Specifically, we split the dataset into 15,312 samples for training and 796 samples for validation. Following the protocols in~\cite{chen2025transanimate, wang2025transpixeler, yin2025qwen}, we employ several evaluation metrics to assess the performance of \textbf{TransText}. For quantitative evaluation, we adopt the \emph{Frechet Video Distance} (FVD)~\cite{unterthiner2018towards} to measure the quality of RGB generation, and the \emph{soft $\alpha$ mean Intersection over Union} (Soft $\alpha$-mIoU) to evaluate $\alpha$ channel prediction. Furthermore, we introduce the \emph{RGBA alignment score}, which assesses the motion alignment consistency between the RGB and Alpha channels, computed using optical flow. In addition to these objective metrics, we conduct a user study to evaluate the perceptual motion quality of the generated RGBA videos. Refer to the Appendix for more details.

\noindent{\textbf{Implementation Details}}. We implement our \textbf{TransText} model based on the Wan-I2V architecture~\cite{wan2025wan}, which contains 14B parameters. We strictly adhere to the original configurations of the Wan model regarding frame resolution and temporal sampling step for all RGB and Alpha videos.  Given that most visual effects exhibit fade-in and fade-out characteristics, we select the middle frame as the reference image to ensure the visual completeness of the glyph, and set $\beta = 5$ for trimap generation. During training, we perform full fine-tuning for $\mathcal{V}$ while keeping the VAE, image encoder, and text encoder frozen. We employ a cosine learning rate scheduler with an initial learning rate of $1.0 \times 10^{-5}$, and use the AdamW optimizer~\cite{loshchilov2019decoupled} with a weight decay of $1.0 \times 10^{-2}$. The training is conducted for $5$ epochs with a batch size of $256$. The balancing loss weight is set to $\lambda = 0.3$. For diffusion-related settings, we set the total number of sampling timesteps to $1000$. During inference, the number of sampling steps is set to $50$, with a classifier-free guidance scale of $5$. The training (inference) is performed on NVIDIA A100 (H100) GPUs.

% \begin{table}[htbp]
% \centering
% \caption{Distribution of visual effects in the dataset, showing sample counts for \emph{training} (Train) and \emph{validation} (Val) splits.}
% \label{tab:effect_distribution}
% \begin{tabular}{|c|l|r|r|}
% \hline
% \textbf{Index} & \textbf{Effect} & \textbf{Train} & \textbf{Val} \\
% \hline
% 1 & Glitch & 2,945 & 138 \\
% 2 & from the sand & 2,146 & 119 \\
% 3 & flickering green & 2,103 & 109 \\
% 4 & Letters are collected & 1,950 & 100 \\
% 5 & small squares & 1,712 & 92 \\
% 6 & lightning strikes & 1,624 & 82 \\
% 7 & Snow falls & 1,193 & 69 \\
% 8 & dark fire & 884 & 52 \\
% 9 & Blue lights & 755 & 35 \\
% \hline
% \multicolumn{2}{|c|}{\textbf{Total}} & \textbf{15,312} & \textbf{796} \\
% \hline
% \end{tabular}
% \end{table}

% 

% Since the generated videos has no accurate frame-wise aligned ground-truth for alpha channel, we merely evaluate its single reference-image aligned 

\subsection{Comparison with Existing Alternatives}\label{sec_experiment_sota}

\begin{figure*}[t]
    \centering
    \includegraphics[width=0.98\linewidth]{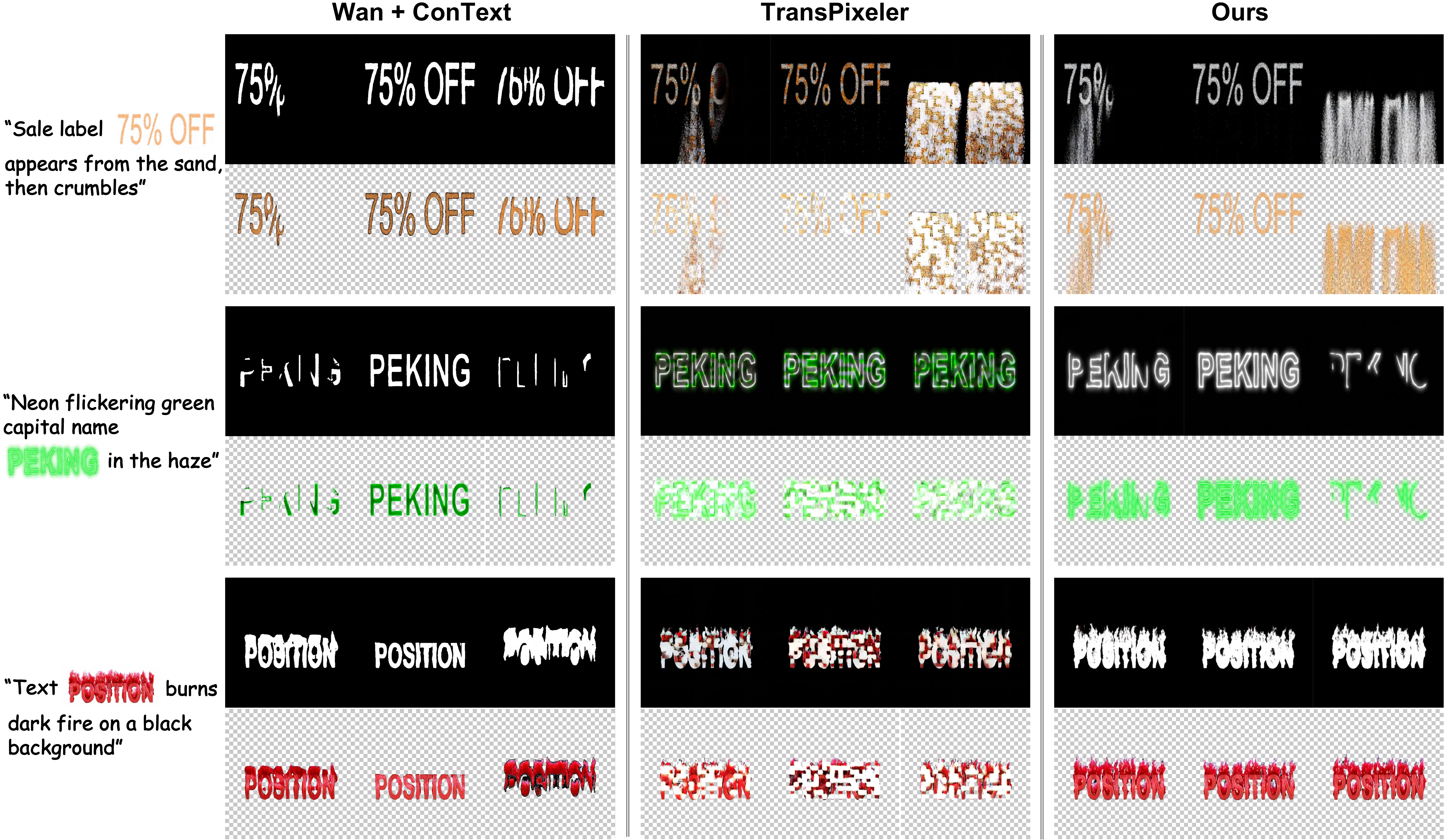} 
\caption{\textbf{Comparison with various I2V generalist models on our text animation dataset.} Each example shows the high-fidelity, motion-consistent RGB video (top row) and the corresponding transparent video (bottom row) generated by \textbf{TransText}. Zoom in for better details.}
    \label{fig_exp_vis_main}
\end{figure*}

\noindent{\textbf{Comparison with Prediction-based Baseline.}} One straightforward approach to RGBA video generation is a \textit{generation-then-prediction} pipeline: first generating an RGB video and then estimating its transparency channel using a video matting model~\cite{wang2025transpixeler}. To implement this baseline, we fine-tune Wan~\cite{wan2025wan}, a standard I2V generalist model and apply a glyph-aware matting model~\cite{zhang2025context} for frame-wise $\alpha$ prediction on the generated video. As shown in Table~\ref{app_tab_general_sota}, while this two-stage pipeline can produce a plausible transparent effect, it suffers from suboptimal motion-aware alignment between RGB and Alpha channels due to its decoupled design—unlike joint generation approaches. This limitation leads to degraded performance in motion-coherent Alpha estimation. Figure~\ref{fig_exp_vis_main} visually confirms the shortcomings of prediction-based pipeline. In contrast, our method achieves superior performance in motion-consistent RGBA synthesis, particularly in accurately capturing dynamic Alpha transitions.

\noindent{\textbf{Comparison with RGBA Baseline.}} We also compare our method with TransPixeler~\cite{wang2025transpixeler}, a state-of-the-art VAE-training-free pipeline. Originally targeting general object-centric domains under T2V setting, TransPixeler achieves joint RGB-and-Alpha learning via temporal concatenation. To adapt it to our I2V glyph generation task, we follow the open-sourced ``T2V Wan-version'' of TransPixeler and re-implement it within an I2V framework using the same training protocol and hyper-parameters as our method. 

As shown in Table~\ref{app_tab_general_sota}, however, this adapted variant exhibits severe mixing artifacts between RGB and Alpha channels, leading to extremely poor transparency prediction. The core issue stems from its mechanism for distinguishing the Alpha stream: rather than assigning separate positional embeddings, TransPixeler shares the \emph{identical positional embeddings} between RGB and Alpha, and relies solely on a single zero-initialized learnable embedding token appended to the Alpha input to differentiate the two modalities. While this design aims to preserve motion alignment by enforcing shared spatio-temporal structure, it renders the separation between RGB and Alpha highly dependent on the stability of that auxiliary learnable token once it is learnt as trivial solution. In practice, we found this token fails to consistently disentangle appearance from transparency in Wan, causing the model to treat RGB and Alpha as nearly indistinguishable signals during training. Consequently, $\alpha$ collapses into a mixture of color and opacity information, manifesting as structured blending artifacts. Furthermore, the temporal concatenation strategy exacerbates this problem by introducing reference frame selection on either RGB or Alpha pattern (as discussed in Section~\ref{sec_method}), which further encourages color leakage into the $\alpha$ predictions. Figure~\ref{fig_exp_vis_main} clearly illustrates this $\alpha$-inferior pattern, where predicted transparencies are contaminated by RGB textures.

In contrast, our method achieves robust and disentangled transparent video generation while maintaining strong motion alignment,
demonstrating the superiority of our spatially aware, alignment-preserving architecture over naive token-based modality separation.

\noindent{\textbf{Efficiency Analysis}}. Table~\ref{app_tab_general_sota} reports the training time per epoch and inference time per video. Reasonably, the generation-then-prediction pipeline is the fastest, owing to its decoupled design that leverages an RGB-only I2V model and a lightweight matting head. Regarding joint modeling approaches, our TransText exhibits comparable efficiency to TransPixeler, with both incurring higher computational overhead than the baseline. However, despite the similar cost, TransText significantly outperforms TransPixeler in terms of generation quality and alignment, making it the superior choice for high-fidelity RGBA video synthesis.

% Our \textbf{TransText} strikes a favorable balance it achieves the best generation and alignment performance while maintaining

% Our \textbf{TransText} strikes a favorable balance: it achieves the best generation and alignment performance while maintaining moderate training cost (69.72 min/epoch)—only $20$\% slower than the baseline—and reasonable inference speed (10.11 min/video), making it practical for high-quality RGBA video synthesis.

\begin{table*}[t]
    \centering
    \caption{\textbf{Comparison with existing RGBA-based methods on our validation set.} \textbf{TransText} outperforms all baselines in  generation quality with promising efficiency.
\textsuperscript{*}Fine-tuned on our training data.
\textsuperscript{\dag}Scores derived from a user study. \textbf{Bold} indicates the best performance, and \underline{underlined} indicates the second best. }
    \resizebox{1.0\textwidth}{!}{
    \begin{tabular}{l cccc c cccc c cccc}
        \toprule[1pt]
        \multirow{2}{*}{Method} & \multicolumn{4}{c}{Generation Metric} & & \multicolumn{4}{c}{Alignment \& Quality Score} && \multicolumn{4}{c}{Time Efficiency}\\
        \cmidrule(lr){2-5} \cmidrule(lr){7-10} \cmidrule(lr){12-15}
        & \multicolumn{2}{c}{FVD$\downarrow$} & \multicolumn{2}{c}{Soft $\alpha$-mIoU$\uparrow$} & & \multicolumn{2}{c}{RGBA Alignment (\%) $\uparrow$}  & \multicolumn{2}{c}{Motion Quality\textsuperscript{\dag}$\uparrow$} & & \multicolumn{2}{c}{Training$\downarrow$} & \multicolumn{2}{c}{Inference$\downarrow$} \\
        % & PSNR  & MSE  & SCUT-Syn & AGE$\downarrow$ & pEPS$\downarrow$ & pCEPs$\downarrow$ & HierText  & TotalText & FST & TextSeg \\
        \midrule
        {Wan\textsuperscript{*} + ConText~\cite{zhang2025context}}  & \multicolumn{2}{c}{\underline{480.92}}& \multicolumn{2}{c}{\underline{57.24}}& & \multicolumn{2}{c}{13.51} & \multicolumn{2}{c}{29.55} &&\multicolumn{2}{c}{\textbf{53.44}} & \multicolumn{2}{c}{\textbf{7.92}} \\
        {Transpixeler\textsuperscript{*}~\cite{wang2025transpixeler}}  & \multicolumn{2}{c}{673.11}& \multicolumn{2}{c}{43.76}& & \multicolumn{2}{c}{\underline{83.01}} & \multicolumn{2}{c}{\underline{64.52}}&& \multicolumn{2}{c}{72.31} &\multicolumn{2}{c}{11.25} \\
        \midrule
        {\textbf{TransText}}  & \multicolumn{2}{c}{\textbf{123.48}}&\multicolumn{2}{c}{\textbf{79.90}}& & \multicolumn{2}{c}{\textbf{87.31}} & \multicolumn{2}{c}{\textbf{75.94}}&&\multicolumn{2}{c}{\underline{69.72}} &\multicolumn{2}{c}{\underline{10.11}}\\
         \bottomrule[1pt]
    \end{tabular}
    }
    \label{app_tab_general_sota}
\end{table*}

\subsection{Ablation Studies}\label{sec_ablation_study}

\begin{figure}[t]
    \centering
    % --- 左侧：表格 ---
    \begin{minipage}{0.58\linewidth}
        \centering
        \captionof{table}{\textbf{Ablation studies on each designed module.} \textbf{Bold} indicates the best performance, and \underline{underlined} indicates the second best.}
        \label{app_tab_ablation}
        \resizebox{\linewidth}{!}{
        \begin{tabular}{l c c c}
            \toprule[1pt]
            {Method} & FVD$\downarrow$ & $\alpha$-mIoU$\uparrow$ & RGBA Alignment$\uparrow$ \\
            \midrule
            Baseline ($w$-wise Concatenation) &367.64&65.95&85.35 \\
            \rowcolor{gray!20} \textit{Concatenation Manner} &  & & \\
            $h$-wise Concatenation &369.52&65.16&84.13 \\
            $t$-wise Concatenation &451.35 &61.28 & 73.81\\
            \rowcolor{gray!20} \textit{Conditional Color Alignment (ColorAlign)} &  & & \\
            Baseline + trimap-based reference & 199.67  & 74.84& \textbf{86.15}\\
             \rowcolor{gray!20} \textit{Alpha Rectification (PixelSup)} &  & & \\
             Baseline + $\mathcal{L}_{rec}$ ($\lambda = 0.1$)  & 215.71 & 72.25  &85.02 \\
             Baseline + $\mathcal{L}_{rec}$ ($\lambda = 0.3$)  & \textbf{143.62} & \textbf{77.25}  &\underline{85.88} \\
             Baseline + $\mathcal{L}_{rec}$ ($\lambda = 0.5$)  & \underline{155.91} & \underline{77.07}  &85.32 \\
             Baseline + $\mathcal{L}_{rec}$ ($\lambda = 0.8$)  & 177.97 & 76.15  &85.57 \\
             \bottomrule[1pt]
        \end{tabular}
        }
    \end{minipage}
    \hfill % 在两个 minipage 之间增加弹性间距
    % --- 右侧：图片 ---
    \begin{minipage}{0.38\linewidth}
        \centering
        \includegraphics[width=\linewidth]{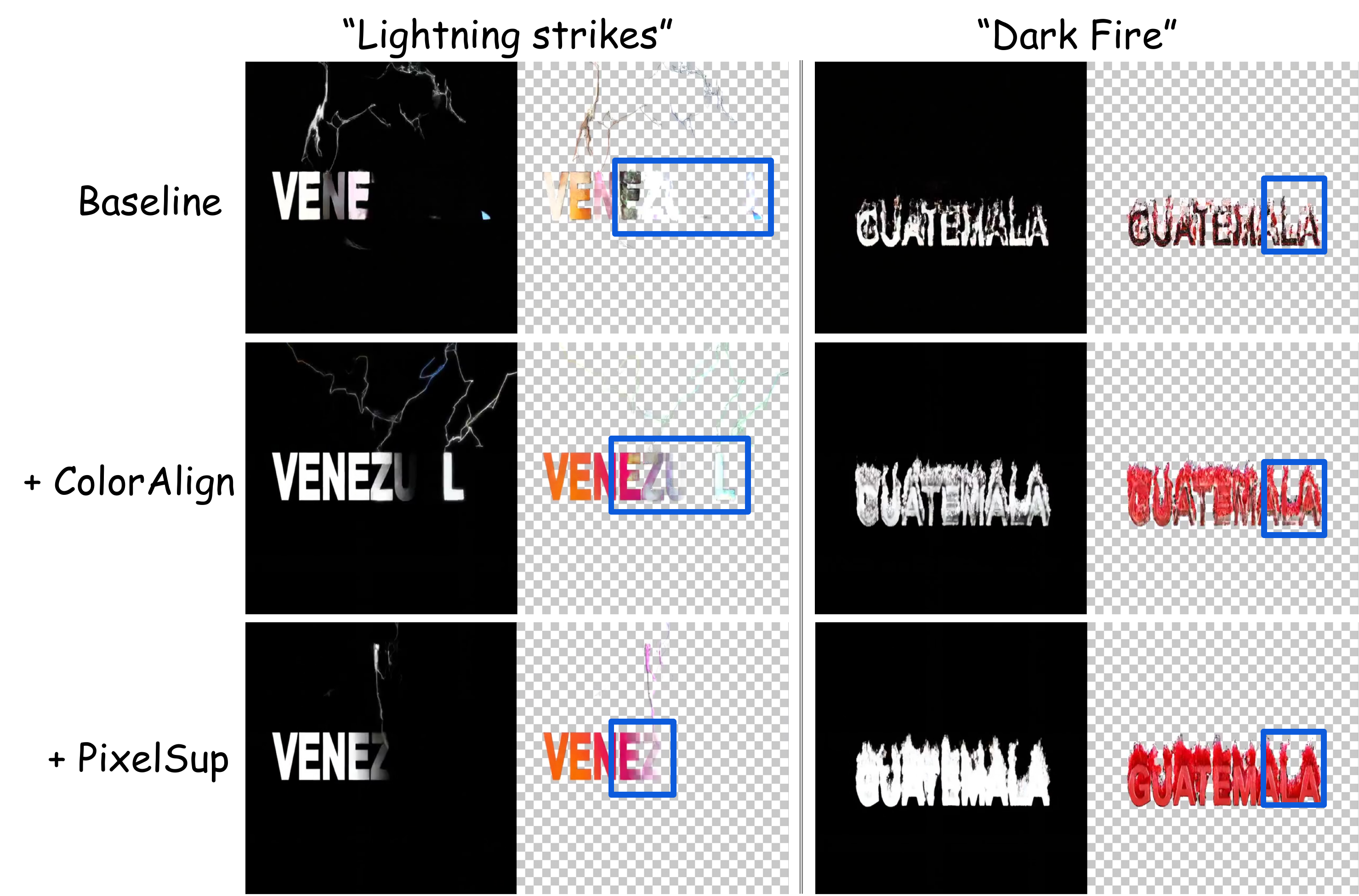} 
        \caption{\textbf{Visual comparison of ColorAlign and PixelSup.} Both enhance $\alpha$ learning.}
        \label{fig_alpha_fine}
    \end{minipage}
\end{figure}

\noindent{\textbf{Spatial \emph{vs} Temporal Concatenation}}.  As shown in Table~\ref{app_tab_ablation}, both spatial concatenation variants, i.e., width-wise ($w$-wise) and height-wise ($h$-wise), consistently outperform temporal concatenation ($t$-wise) across all metrics. Notably, the $t$-wise variant suffers from a significant drop in FVD and RGBA alignment, indicating degraded motion coherence and severe channel mixing. This degradation arises from two intertwined limitations of temporal concatenation. First, as discussed in Section~\ref{sec_method}, during training and inference, the reference image  must be selected as either the RGB image or the trimap. This forces the model to hallucinate the missing modality, leading to ambiguous cross-channel signals and structured mixing artifacts. Second, over longer video sequences, temporal error propagation amplifies this instability~\cite{waseem2025video}, causing the $\alpha$ channel to increasingly absorb RGB appearance cues. Thus, similar mixing patterns are also observed in TransPixeler~\cite{wang2025transpixeler}. In contrast, spatial concatenation embeds RGB and $\alpha$ features within the same spatial grid at each timestep, preserving per-frame structural correspondence and enabling explicit, stable co-learning of both modalities. The comparable performance between $w$-wise and $h$-wise further highlights the effectiveness of spatial alignment for robust RGBA generation.

% Misaligned time concatenation
% genads_models/tree/feistar/output/ia2v/wan21_i2v_test_480p/2146881485/10/r/1663560

\noindent{\textbf{Fine-grained Alpha Learning}}. Table~\ref{app_tab_ablation} studies the effectiveness of the designed two $\alpha$-based fine-grained learning modules, \textit{i.e.}, trimap adaptation and latent-level regularization $\mathcal{L}_\mathrm{rec}$. Notably, both components contribute positively to RGBA video generation, as evidenced by consistent improvements across all metrics compared to the baseline. Notably, the latent-level rectification via $\mathcal{L}_\mathrm{rec}$ yields the most substantial gain, particularly in motion fidelity and $\alpha$ accuracy. Qualitative results in Figure~\ref{fig_alpha_fine} further underscores that videos produced with $\mathcal{L}_\mathrm{rec}$ exhibit cleaner transparency boundaries and reduced RGB-Alpha mixing artifacts.

Additionally, we further analyze the sensitivity of the latent-level supervision $\mathcal{L}_\mathrm{rec}$ to its weighting coefficient $\lambda$. As $\lambda$ increases from 0.1 to 0.3, both FVD ($215.71 \rightarrow$  $143.62$) and $\alpha$-mIoU ($72.25$ $\rightarrow$ $77.25$) improve significantly, confirming that stronger latent-wise supervision enhances Alpha reconstruction. However, pushing $\lambda$ further  leads to diminishing returns, suggesting that excessive emphasis on Alpha rectification can inadvertently compromise the joint appearance-transparency balance.

% \section{Limitations}
% Our work focuses specifically on I2V transparent glyph animation. While this enables precise control for typographic applications, it may fail to generalize to other paradigms in natural-image domains without further task-general finetuning. Moreover, due to the labor-intensive nature of high-quality RGBA annotation—particularly for transparency dynamics—the scale of our training data remains limited. We believe that future efforts in scalable annotation pipelines for transparent layer significantly advance transparent video generation in this underexplored domain.

\section{Conclusion}

This paper presents the first study on adapting the standard \emph{image-to-video} (I2V) diffusion paradigm to layer-aware transparent glyph animation. In contrast to prevailing VAE-training-based frameworks that decouple RGB and $\alpha$ into separate appearance and transparency modeling, we introduce a unified, joint RGB-and-Alpha representation. By explicitly representing the $\alpha$ channel within the RGB space, our method eliminates the need for resource-intensive latent encoder reconstruction, thereby preserving the robust generative priors of the pre-trained backbone. Based on this \emph{Alpha-as-RGB} paradigm, we employ a direct spatial concatenation strategy. Unlike temporal alignment approaches that distort sequence length, our spatial design explicitly aligns RGB and $\alpha$ learning throughout the diffusion process, enabling coherent motion and precise opacity control. Furthermore, we introduce $\alpha$-oriented modifications through a latent-wise regularization term that explicitly constrains the generation of opacity. This term serves to eliminate edge artifacts and enforce strict boundary alignment, facilitating fine-grained transparency modeling crucial for high-fidelity RGBA video generation. Extensive experiments demonstrate the effectiveness and superiority of our framework in producing temporally smooth, visually plausible, and structurally accurate transparent glyph animations. We hope this pioneering work will inspire further research into transparent video generation, particularly in the underexplored domain of typographic and glyph-based animation.

\clearpage
\newpage
\bibliographystyle{assets/plainnat}
\bibliography{paper}

\clearpage
\newpage

\clearpage % 强制换页

\begin{figure*}[t!] % 双栏模版一定要用 figure*，单栏用 figure
    \centering
    \vfill % 顶部填充，把图往下顶
    
    % 插入你的图片
    \includegraphics[width=0.84\textwidth,height=0.97\textheight]{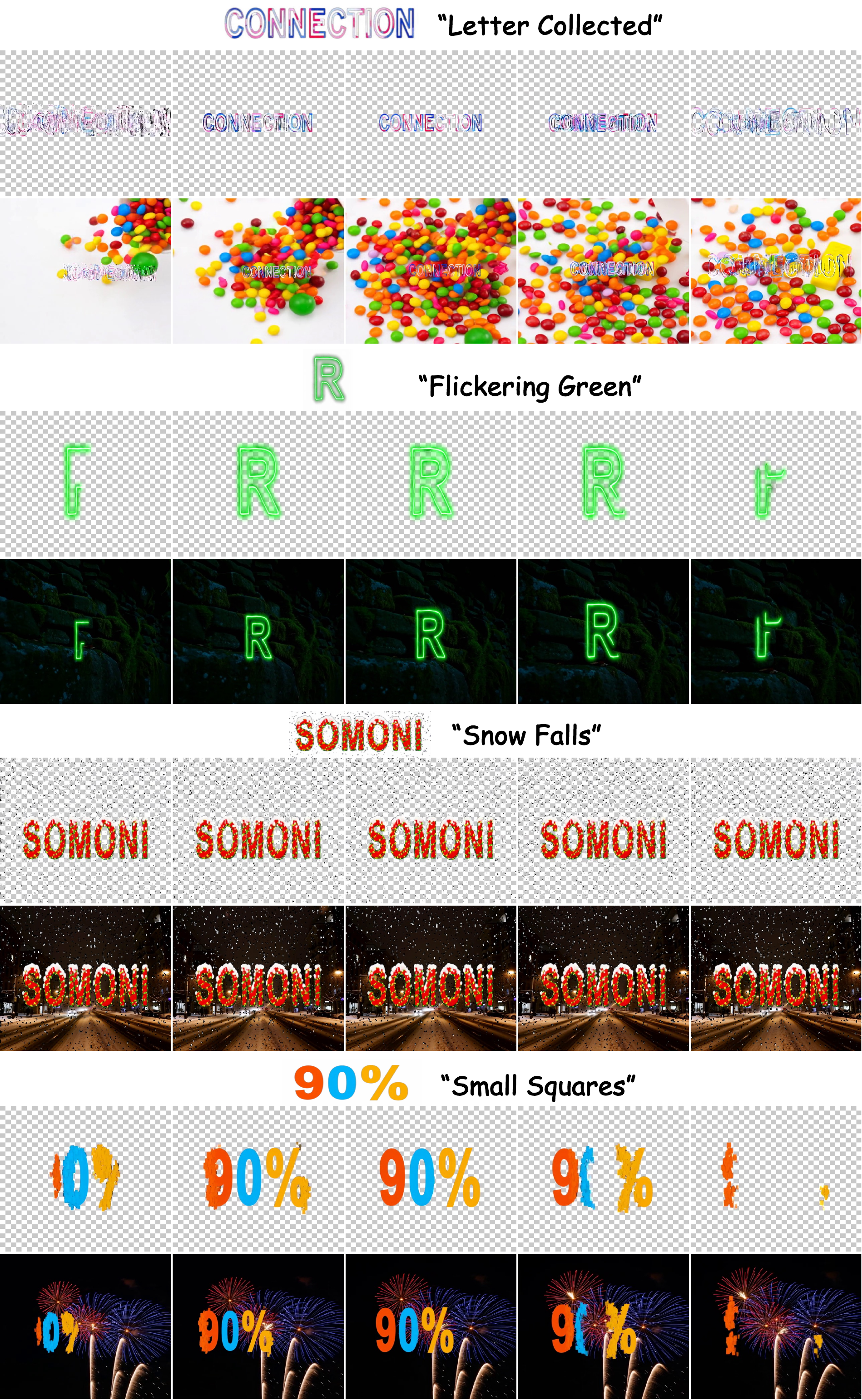}
    \caption{\textbf{Qualitative results on background compositing.} Visual effects include: \textit{Letter Collection}, \textit{Flickering Green}, \textit{Snow Falls}, and \textit{Small Squares}.}
    \label{fig_fo_1}
    \vfill % 底部填充，把图往上顶
\end{figure*}

\clearpage % 强制换页，恢复正文

\clearpage % 强制换页

\begin{figure*}[t!] % 双栏模版一定要用 figure*，单栏用 figure
    \centering
    \vfill % 顶部填充，把图往下顶
    
    % 插入你的图片
    \includegraphics[width=0.84\textwidth,height=0.95\textheight]{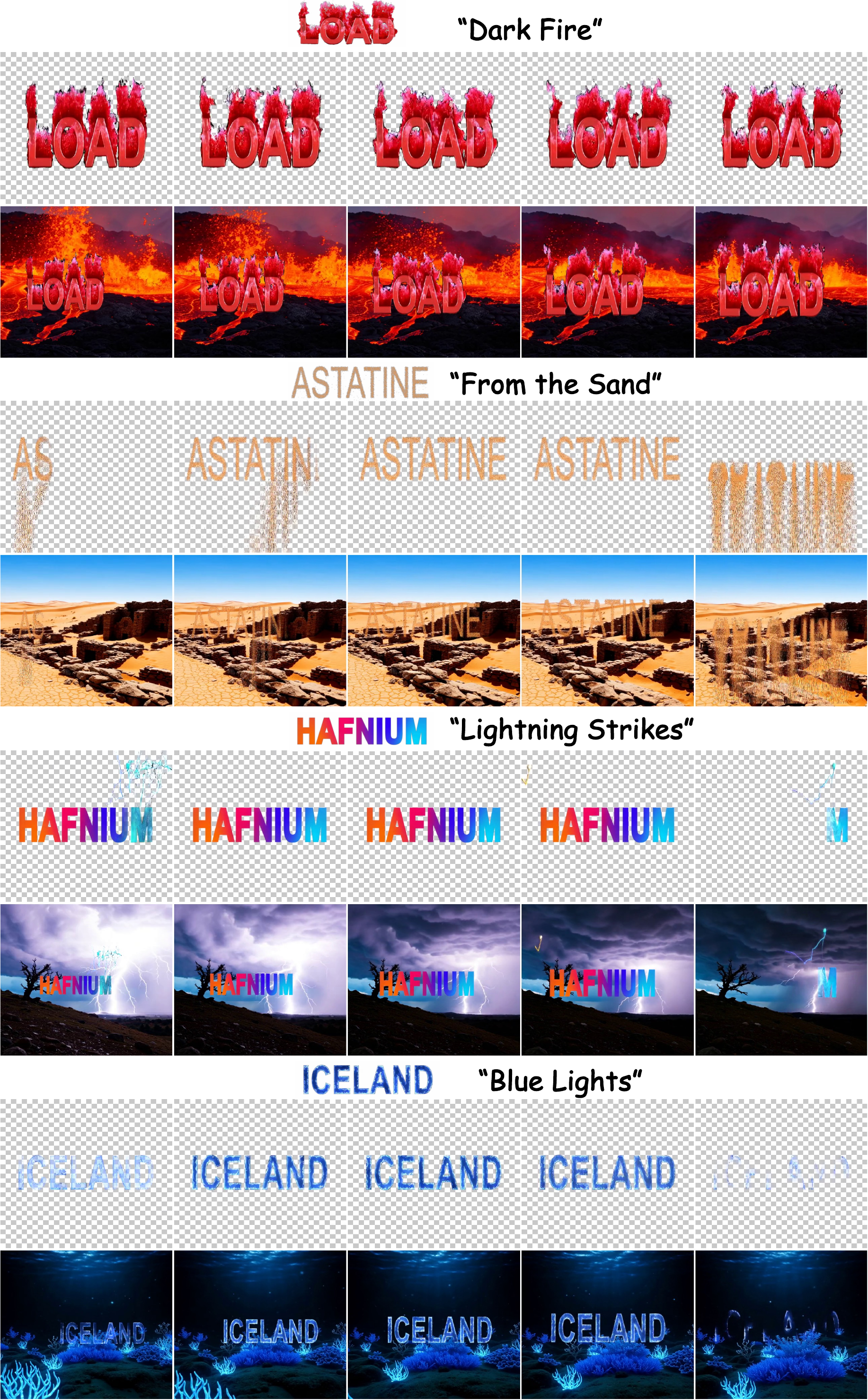}
    
    \caption{\textbf{Qualitative results on background compositing.} Visual effects include: \textit{Dark Fire}, \textit{From the Sand}, \textit{Lightning Strikes}, and \textit{Blue Lights}.}
    \label{fig_fo_2}
    
    \vfill % 底部填充，把图往上顶
\end{figure*}

\clearpage % 强制换页，恢复正文

\beginappendix
% \newpage
% \appendix
\section{Dataset Introduction}\label{appendix_dataset}
As described in Section~\ref{sec_exp_setting}, we utilize a self-constructed RGBA text animation benchmark for all our experiments. This dataset comprises eight distinct visual effects, with their statistical distribution presented in Table~\ref{tab_effect_distribution}. Similar to segmentation tasks~\cite{zhang2021complementary,li2023monte,ma2023diffusionseg,zhou2024image,zhang2023uncovering,ma2023attrseg,yang2024multi,liu2023zero,liu2024audio,liu2025lamra,wang2023crack,li2023ddaug,chen2024probabilistic,zhang2025decouple,zhang2025g4seg,mai2023exploit}, we also take some off-the-shelf segmentation tools to help refine the Alpha channel. To further illustrate the diversity among these effects, we provide visualizations of each effect applied to the original RGB video in Figure~\ref{fig_shutterstock}, which clearly highlights their differences. Additionally, to ensure precise alignment between the reference text and video, we employ QwenVL~\cite{bai2025qwen3vltechnicalreport} to refine the captions.

\setcounter{table}{0}
\begin{table}[ht]
    \centering
    \caption{\textbf{Distribution of 8 visual effects in the dataset.} Below shows the sample counts for \emph{training} (Train) and \emph{validation} (Val) splits.}
    \label{tab_effect_distribution}
    % \resizebox{0.7\linewidth}{!}{
    \begin{tabular}{c l r r}
        \toprule[1pt]
        \textbf{ID} & \textbf{Effect} & \textbf{Train} & \textbf{Val} \\
        \midrule
        1 & From the sand & 2,515 & 137 \\
        2 & Flickering Green & 2,471 & 127 \\
        3 & Letters Collected & 2,318 & 117 \\
        4 & Small Squares & 2,080 & 109 \\
        5 & Lightning Strikes & 1,992 & 99 \\
        6 & Snow Falls & 1,561 & 86 \\
        7 & Dark Fire & 1,252 & 69 \\
        8 & Blue Lights & 1,123 & 52 \\
        \midrule
        & \textbf{Total} & \textbf{15,312} & \textbf{796} \\
        \bottomrule[1pt]
    \end{tabular}
    % }
\end{table}

\setcounter{figure}{0}
\begin{figure*}[t]
    \centering
    \includegraphics[width=1.0\linewidth]{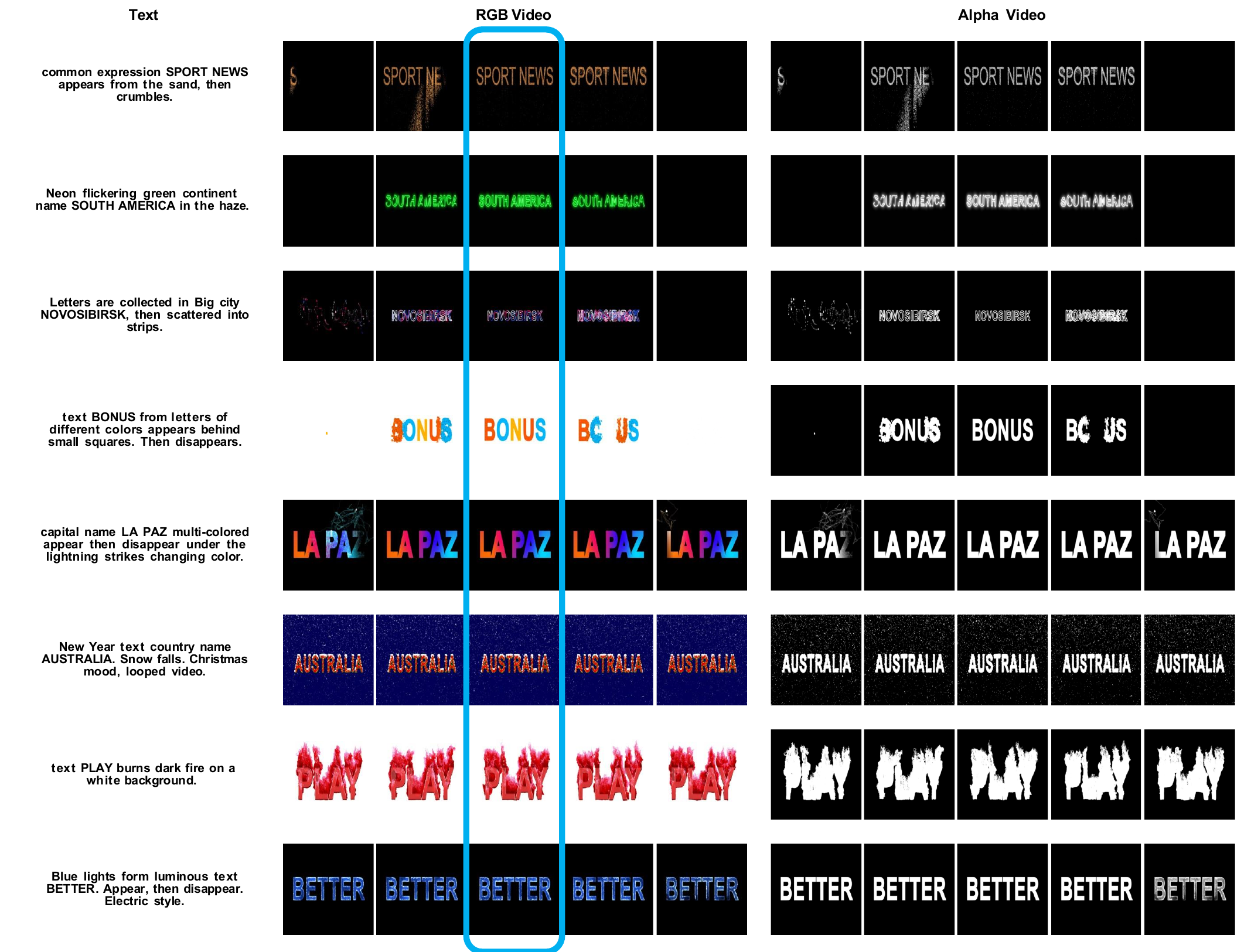} 
    % \vspace{-8mm}
    \caption{\textbf{Visualization of original samples from our glyph animation dataset, with one example shown for each visual effect.} The reference image (the column marked in blue) corresponds to the middle frame of each video clip. Please zoom in for a better view.}
    \label{fig_shutterstock}
\end{figure*}

\section{RGBA Alignment Score}\label{appendix_rgb_alignment}
In Section~\ref{sec_experiment}, we introduce a specifically-designed evaluation metric, RGBA Alignment, to assess the motion alignment between the RGB foreground and the Alpha video. This metric is primarily based on the optical flow similarity between the RGB and Alpha channels. It compares the dense optical flow fields computed from both videos on a frame-by-frame basis, capturing discrepancies in motion direction, magnitude, and spatial alignment. Concretely, for each pair of consecutive frames in both videos, we compute forward optical flows using the Farneback~\cite{farneback2003two} method for efficiency. The resulting flow fields are then evaluated using four complementary metrics:  
\begin{enumerate}
    \item \textbf{End-point Error}~\cite{baker2011database}: Measures pixel-wise displacement deviation.
    \item \textbf{Angular Error}~\cite{butler2012naturalistic}: Assesses directional misalignment.
    \item \textbf{Magnitude Similarity}: Calculated via Pearson correlation of flow magnitudes.
    \item \textbf{Directional Consistency}: Measures cosine similarity of normalized flow vectors.
\end{enumerate}
These metrics are integrated into a unified score ranging from $0$ to $100$, where higher values indicate better motion alignment—i.e., less motion discrepancy between the RGB and Alpha layers. For large-scale evaluation, we employ the Farneback method due to its computational efficiency, while RAFT is reserved for qualitative analysis on selected samples. The detailed calculation procedure is provided in Algorithm~\ref{alg:rgba_align}.

\begin{algorithm}[htbp]
\caption{RGBA Alignment Score Computation}
\label{alg:rgba_align}
\begin{algorithmic}[1]
\REQUIRE $V_{\text{rgb}}$: RGB video frames, $V_{\alpha}$: Alpha video frames

\STATE Compute optical flow for $V_{\text{rgb}}$: $F_{\text{rgb}} = \text{Flow}(V_{\text{rgb}})$
\STATE Compute optical flow for $V_{\alpha}$: $F_{\alpha} = \text{Flow}(V_{\alpha})$

\STATE Calculate EPE: $epe = \frac{1}{H \cdot W \cdot T} \sum \| F_{\text{rgb}} - F_{\alpha} \|_2$
\STATE Calculate Angular Error: $\theta = \angle(F_{\text{rgb}}, F_{\alpha})$, averaged over pixels
\STATE Calculate Magnitude Similarity: $\rho = \text{PearsonCorr}(\|F_{\text{rgb}}\|, \|F_{\alpha}\|)$
\STATE Calculate Directional Consistency: $\cos\phi = \frac{F_{\text{rgb}} \cdot F_{\alpha}}{\|F_{\text{rgb}}\| \cdot \|F_{\alpha}\|}$, averaged over valid regions

\STATE Normalize scores:
\begin{itemize}
    \item $s_{\text{epe}} = \exp(-epe / 10.0)$
    \item $s_{\text{angle}} = \exp(-\theta / 45.0)$
    \item $s_{\text{mag}} = (\rho + 1)/2$
    \item $s_{\text{dir}} = (\cos\phi + 1)/2$
\end{itemize}

\STATE Aggregate: $\text{Final Score} = 0.25 \cdot (s_{\text{epe}} + s_{\text{angle}} + s_{\text{mag}} + s_{\text{dir}})$

\end{algorithmic}
\end{algorithm}

\section{Motion Quality}
We also introduce a human-level evaluation metric for generative video quality in \textbf{TransText}, implemented via a user study involving 54 participants. Participants were asked to assess (1) which animation best aligns with the given reference text and image, and (2) whether the motion quality conforms to human aesthetic preferences.

\section{Alpha Attention} 
\begin{figure}[t]
    \centering
    % --- 左侧放图片 ---
    \begin{minipage}[b]{0.48\linewidth}
        \centering
        \includegraphics[width=1.0\linewidth]{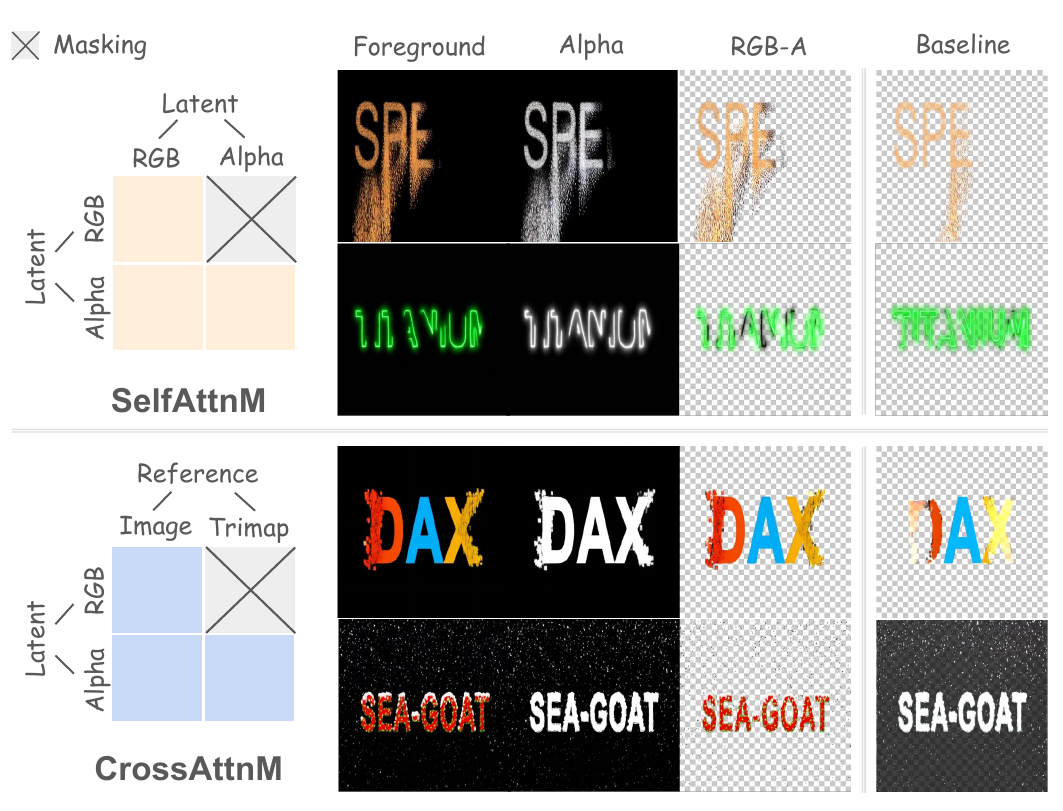}
        \caption{\textbf{Visualized results of different $\alpha$-oriented attention masking mechanisms.} Clearly, CrossAttnM improves transparency estimation and video quality, while SelfAttnM harms RGB–$\alpha$ alignment and generation fidelity.}
        \label{fig_attention}
    \end{minipage}
    \hfill % 在两个 minipage 之间填满空间
    % --- 右侧放表格 ---
    \begin{minipage}[b]{0.48\linewidth}
        \centering
        \captionof{table}{\textbf{Analysis on attention rectification.} The table reports the performance of adding rectification mechanisms to Self-Attention (SelfAttnM) and Cross-Attention (CrossAttnM) layers. \textbf{Bold} indicates best, \underline{underlined} indicates second best.}
        \label{app_tab_ablation_attn}
        \resizebox{\linewidth}{!}{
            \begin{tabular}{l c c c}
                \toprule[1pt]
                Method & FVD$\downarrow$ & $\alpha$-mIoU$\uparrow$ & RGBA Alignment$\uparrow$ \\
                \midrule
                Baseline & \underline{367.64} & \underline{65.95} & \textbf{85.35} \\
                Baseline + SelfAttnM & 451.27 & 65.87 & 77.72 \\
                Baseline + CrossAttnM & \textbf{274.73} & \textbf{72.61} & \underline{84.99} \\
                \bottomrule[1pt]
            \end{tabular}
        }
    \end{minipage}
\end{figure}

To enhance both 
 $\alpha$-matte and RGB generation, \cite{wang2025transpixeler} has shown that $\alpha$-orient  attention masking plays a crucial role in latent fusion for T2V synthesis. 
To further study the role of transparency-aware attention in our \textbf{TransText} framework, we propose two complementary masking mechanisms: \emph{self-attention masking} (SelfAttnM) and \emph{cross-attention masking} (CrossAttnM). SelfAttnM masks the attention from $\alpha$ to the RGB latent during self-attention, aiming to isolate appearance learning from transparency-driven influence. In contrast, CrossAttnM explicitly masks the attention from the input text to 
$\alpha$ in cross-attention, preventing semantic cues from directly interfering with transparency estimation and thereby encouraging 
$\alpha$ to rely more on structural priors from the reference trimap. As shown in Table~\ref{app_tab_ablation_attn}, SelfAttnM degrades overall performance—hurting video quality and weakening RGBA alignment—without improving  estimation, suggesting that allowing 
$\alpha$ to inform RGB representation is beneficial for coherent generation. On the other hand, CrossAttnM consistently improves both transparency accuracy and generation fidelity, highlighting the importance of $\alpha$-aware attention for RGBA video generation.

\clearpage

\end{document}